\documentclass[runningheads]{llncs}
\usepackage[T1]{fontenc}
\usepackage{algorithmic}
\usepackage{textcomp}
\usepackage{tabularx}
\usepackage{makecell}
\usepackage{stfloats}
\usepackage{algorithmic}

\usepackage{tabularx}
\usepackage{array}
\usepackage{makecell}
\usepackage{booktabs}
\usepackage{amsmath}
\usepackage{graphicx}
\begin{document}
\title{FMNV: A Dataset of Media-Published News Videos for Fake News Detection}
%
%
\author{Yihao Wang, Zhong Qian, Peifeng Li}
%

%
\institute{}
\maketitle              
\begin{abstract}
News media, particularly video-based platforms, have become deeply embed-ded in daily life, concurrently amplifying the risks of misinformation dissem-ination. Consequently, multimodal fake news detection has garnered signifi-cant research attention. However, existing datasets predominantly comprise user-generated videos characterized by crude editing and limited public en-gagement, whereas professionally crafted fake news videos disseminated by media outlets—often politically or virally motivated—pose substantially greater societal harm. To address this gap, we construct FMNV, a novel da-taset exclusively composed of news videos published by media organizations. Through empirical analysis of existing datasets and our curated collection, we categorize fake news videos into four distinct types. Building upon this taxonomy, we employ Large Language Models (LLMs) to automatically generate deceptive content by manipulating authentic media-published news videos. Furthermore, we propose FMNVD, a baseline model featuring a dual-stream architecture that integrates spatio-temporal motion features from a 3D ResNeXt-101 backbone and static visual semantics from CLIP. The two streams are fused via an attention-based mechanism, while co-attention modules refine the visual, textual, and audio features for effective multi-modal aggregation. Comparative experiments demonstrate both the generali-zation capability of FMNV across multiple baselines and the superior detec-tion efficacy of FMNVD. This work establishes critical benchmarks for de-tecting high-impact fake news in media ecosystems while advancing meth-odologies for cross-modal inconsistency analysis. Our dataset is available in https://github.com/DennisIW/FMNV.

\end{abstract}
\section{Introduction}

The proliferation of video-centric social media platforms has transformed the in-formation landscape, giving rise to increasingly sophisticated forms of multimodal disinformation. Unlike traditional text- or image-based fake news, these video-based narratives integrate textual, visual, and auditory elements to create highly persuasive content~\cite{DBLP:journals/snam/AimeurAB23,DBLP:journals/aiopen/HuWZW22}. Such content exhibits stronger emotional appeal, greater virality, and heightened resistance to detection, making it a growing threat to public opinion and social trust. Recent studies indicate that over 70\% of disinformation incidents on social platforms in 2023 involved coordinated cross-modal manipulation~\cite{DBLP:conf/misdoom/BarberoCKGSI23}, underscoring the urgent need for robust multimodal fake news detection (MFND) systems capable of handling complex and dynamic video data.

Prior work in MFND has primarily focused on static multimodal content, such as image-text combinations, resulting in several influential datasets including FVC-2018~\cite{DBLP:journals/oir/PapadopoulouZPK19}, FakeNewsNet~\cite{DBLP:journals/bigdata/ShuMWLL20}, and MCFEND~\cite{DBLP:conf/www/LiHBW24}. More recent efforts have shifted toward video-based datasets to accommodate the evolving nature of disinformation. Notable among them is FakeSV~\cite{DBLP:conf/aaai/0005BC0SXWC23}, a Chinese short video dataset comprising over 5,000 labeled videos across real, fake, and debunked categories. Building on this, FakeTT~\cite{DBLP:conf/mm/Bu000W024} extended the research to English-language platforms like TikTok, focusing on multimodal inconsistencies. These resources have significantly advanced the field by introducing temporal dynamics and audiovisual complexity into the analysis.

However, these datasets mainly consist of user-generated content, which often lacks wide influence and is easier to verify. In contrast, fake news from mainstream media—driven by political or economic agendas—is more polished, widespread, and impactful, yet remains underrepresented. Additionally, large-scale manual annotation is costly and prone to error, and existing datasets suffer from class imbalance skewed toward real news. To address these issues, we introduce \textbf{F}ake \textbf{M}edia \textbf{N}ews \textbf{V}ideos (\textbf{FMNV}), a dataset of 2,393 professionally produced news videos sourced from verified YouTube and Twitter accounts. To mitigate annotation and imbalance challenges, we employ large language models (LLMs) to generate 1,500 synthetic fake news videos based on empirically defined misinformation categories. Each video features multimodal content—titles, audio, and visual clips—with cross-modal inconsistencies or violations of common sense.

We also propose \textbf{F}ake \textbf{M}edia \textbf{N}ews \textbf{V}ideo \textbf{D}etection (\textbf{FMNVD}), a dual-stream baseline model that integrates CLIP and 3D ResNeXt-101 for visual feature extrac-tion, with attention-based gated fusion and co-attention for cross-modal alignment. Experiments on FMNV validate the effectiveness of our framework in detecting complex multimodal disinformation.

\section{The FMNV Dataset}

In this section, we construct a comprehensive dataset through empirical analysis, real-world data collection, LLM-assisted augmentation, and dataset evaluation. We first analyze common semantic inconsistency patterns in fake news videos to guide annotation. Then, we collect and curate video samples from various sources, aug-ment them using large language models and shot segmentation, and finally conduct quantitative and qualitative analysis to compare our dataset with existing bench-marks in terms of quality and diversity.

\subsection{Empirical Analysis}

Existing communication studies~\cite{zimdars2020fake,wardle2017information} suggest that fake news videos can be classi-fied across multiple dimensions, such as production motives, technical means, and content nature. We analyze both public fake news video datasets and real-world samples. Fake news often misleads through edited footage or emotionally manipula-tive, contradictory content. Since news videos typically contain three modalities—title, video, and audio—we examine the semantic differences among them. When one modality conflicts with the others or defies common sense, the video is more likely to be fake. Based on modality-level inconsistency, we categorize fake news videos into four types:

\begin{figure*} \centering \includegraphics[width=\textwidth]{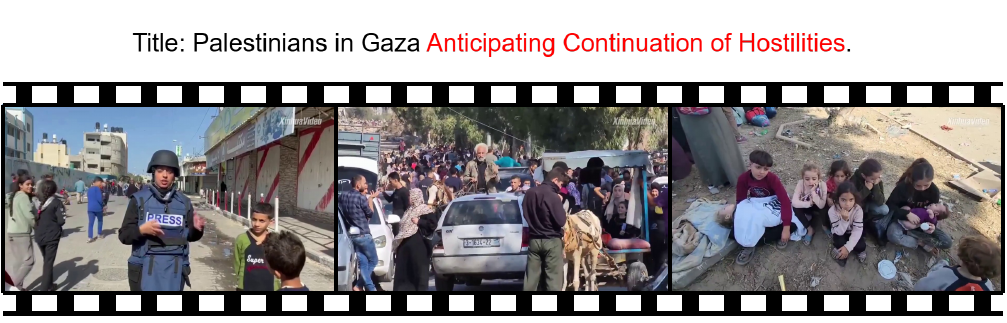} \caption{Example of Contextual Dishonesty: The video depicts children in Gaza longing for peace, while the title claims they anticipate continued conflict.} \label{FNVD} \end{figure*}

\subsubsection{Contextual Dishonesty}

Titles, as the video’s summary, significantly shape audience perception. Malicious title manipulation causes semantic dissonance by omitting context or using emo-tionally charged language. Common tactics include subject-object substitution, ex-aggeration, and removal of qualifiers—misleading readers who engage only with headlines.

\begin{figure*} \centering \includegraphics[width=0.5\textwidth]{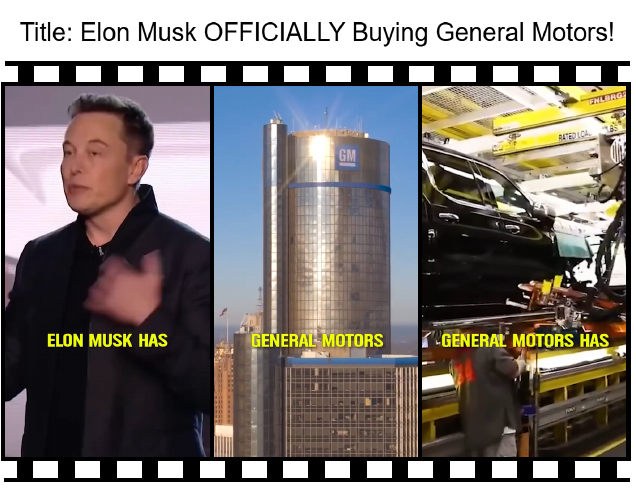} \caption{Example of Cherry-picked Editing: A fabricated claim that Musk is buying GM, stitched together from unrelated clips.} \label{FNVD} \end{figure*}

\subsubsection{Cherry-picked Editing}

Video content is often altered by omitting key scenes or splicing unrelated footage to construct a false narrative. Tactics include removing moderate remarks, excluding peaceful scenes, or combining unrelated events, thereby amplifying bias and mis-leading viewers.

\begin{figure*} \centering \includegraphics[width=\textwidth]{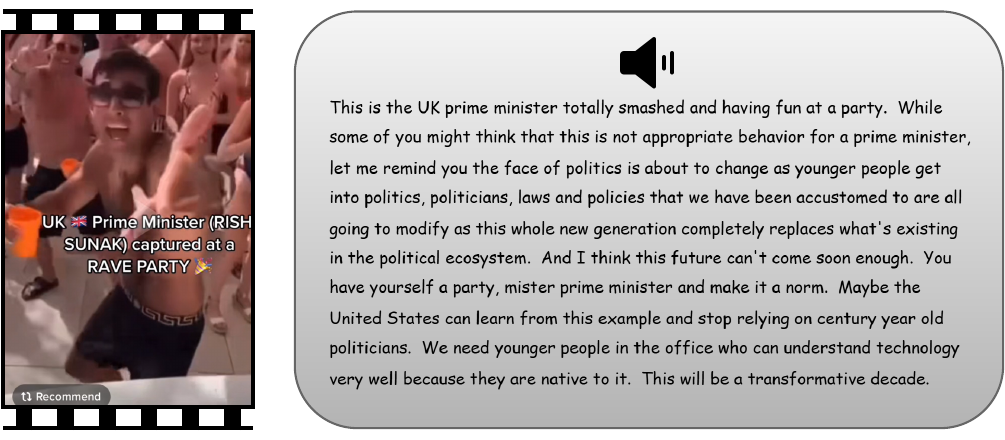} \caption{Example of Synthetic Voiceover: A party video overlaid with unrelated political commentary.} \label{FNVD} \end{figure*}

\subsubsection{Synthetic Voiceover}

These videos pair AI-generated or cloned speech with unrelated visuals. For exam-ple, a fake announcement might play over footage of a harmless event. The mis-match exploits viewers' assumptions of audiovisual coherence, often bypassing de-tection through professional editing.

\begin{figure*}
\centering
\includegraphics[width=0.6\textwidth]{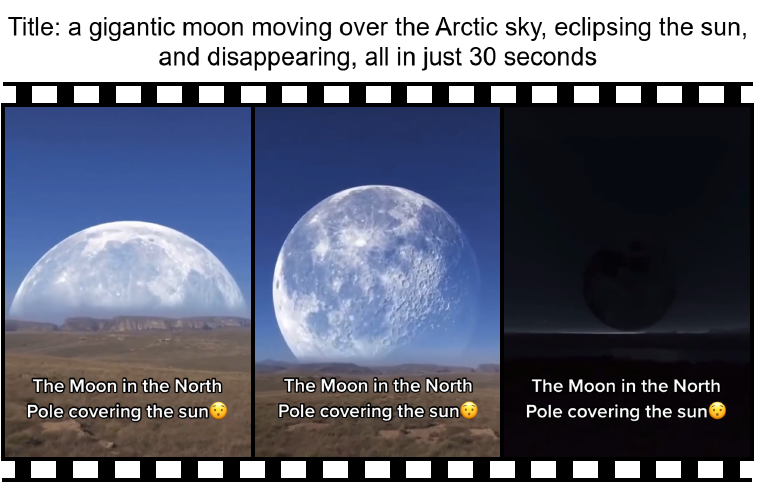}
\caption{An example of Contrived Absurdity fake news video. The video shows a gigantic moon, and the title says it will disappear in 30 seconds, which clearly defies common sense.}
\label{FNVD}
\end{figure*}

\subsubsection{Contrived Absurdity}

Such videos use consistent multimodal content to present absurd claims (e.g., “alien bodies found” or “5-ton dog”), often generated by AI or CGI. Despite surface coher-ence, the narratives defy logic, aiming to attract attention through curiosity and shock.

\subsection{Data Collection}

At present, the existing fake news video detection datasets are collected from the short video platforms Douyin and Tiktok. These videos are mainly published by individual users with a small number of followers. Moreover, the clips of fake news videos are poor and people can easily distinguish the authenticity from the fake ones, so this type of fake news videos is not too harmful. The news video released by the news media is well-produced and will receive more attention, and the news me-dia may also publish fake news videos for political purposes or commercial inter-ests, which is particularly harmful. To solve this problem, we choose to collect news videos posted by news media from Twitter and YouTube. Specifically, we use web crawler technology to collect videos on 12 topics including accidents, epidemics and politics published by 27 news media on Twitter and YouTube in the past five years. In order to minimize noise and improve the quality of the dataset, we manually check and remove the videos with blurred images and too short duration. A total of 2,393 news videos are retained. These news videos contain information in three modals: title, video clip and audio, and the semantic information expressed in all three modals is the same.

\begin{table}
  \centering
  \caption{Examples of prompts and titles that are before and after modification.}
  \label{table:1}
  \setlength{\tabcolsep}{0.3em} 
  \renewcommand{\arraystretch}{1.1} 
  \begin{tabularx}{\textwidth}{
    >{\raggedright\arraybackslash}X@{\hspace{0.5em}}
    >{\raggedright\arraybackslash}X@{\hspace{0.5em}}
    >{\raggedright\arraybackslash}X
  }
    \toprule
    \textbf{Prompt} & \textbf{Original Title} & \textbf{Modified Title} \\
    \midrule
    Title: [title]. Change the meaning of the title from the perspective of object. & A Survey Among Year 11 Students in \textbf{London} Reveals a Majority Would Voluntarily Wear Masks. & A Survey Among Year 11 Students in \textbf{New York} Reveals a Majority Would Voluntarily Wear Masks\\
    \midrule
    Title: [title]. Change the meaning of the title from the perspective of action. & Duke of Edinburgh involved in a road traffic accident, but he is \textbf{not injured}. & Duke of Edinburgh \textbf{injured} in Road Traffic Incident.\\
    \midrule
    Title: [title]. Regenerate a news headline for a completely different event under the same topic & A recent hit-and-run incident on Glebe Road just south of Route 50 near Goodwill. & Multi-Vehicle Pileup Shuts Down I-95 During Morning Rush Hour, Injuries Reported. \\
    \bottomrule
  \end{tabularx}
\end{table}

\subsection{Data Augmentation}

As most collected news videos are from news media, they are predominantly real, resulting in data imbalance. Inspired by Liu et al.~\cite{DBLP:conf/eacl/LiuYS23}, we employ LLMs to generate fake news videos based on real ones. Referring to the four categories of fake news videos in our empirical analysis, we generate corresponding types and ensure similar proportions to those in FakeTT: CD:600, CE:450, SV:300, CA:150.

\textbf{Contextual Dishonesty (CD)}
To introduce semantic mismatches between titles and other modalities, we employ ERNIE 4.0 to rewrite original video titles, ensuring intentional divergence from the actual video content. Prompts are carefully designed to modify specific elements in the title, such as the object, verb phrase, or to generate alternative events under the same general topic. Examples of prompts and corresponding rewritten titles are shown in Table \ref{table:1}. The generated titles are then manually reviewed to remove redun-dancy and refine poorly structured outputs. This process results in CD-type fake news videos, where the title misrepresents the semantic content of the video.

\textbf{Cherry-picked Editing (CE)}
This type introduces semantic inconsistencies through deliberate visual editing. Specifically, we first apply TransNet V2~\cite{Souček_Lokoč_2020} to perform shot segmentation on the original video, dividing it into a series of coherent visual units. From these seg-ments, a portion is evenly removed and the remaining segments are reassembled to create a shortened version of the video. This simulates a cherry-picking process, where critical or context-defining content is selectively omitted. Although the re-sulting video maintains surface-level visual continuity, it loses essential infor-mation, leading to logical incoherence and misalignment with the original title or audio. These manipulated videos are thus categorized as CE-type fake news, charac-terized by cross-modal inconsistencies arising from selective deletion within the visual stream.

\textbf{Synthetic Voiceover (SV)}
Cross-modal contradictions are created in the audio stream by replacing the original narration with unrelated synthesized speech. To achieve this, we use ERNIE 4.0 to generate semantically divergent text while staying loosely on the same topic. The inputs include the video title and image captions of key frames, organized into struc-tured prompts. The generated text is converted to speech using the TTS system VITS, replacing the original audio track. The resulting SV-type fake news videos exhibit clear audio-visual contradictions.

\textbf{Contrived Absurdity (CA)}
This type maintains semantic alignment across modalities but introduces absurdity by violating common sense through exaggerated or illogical claims. Following a similar strategy as in CD, we prompt ERNIE 4.0 to rewrite the title in an exaggerated manner that preserves the original meaning but distorts quantities, causality, or logi-cal reasoning. The prompts avoid altering key entities or events, ensuring the seman-tic core remains intact. The final CA-type videos appear internally consistent but convey implausible or nonsensical information.

Following data augmentation, we end up with a final dataset comprising five cat-egories of 2393 videos.

\begin{table}
  \centering 
  \caption{Comparison of datasets for MFND. DY: Douyin; TT: Tiktok; YT: YouTube; TW: Twitter. The tokens number of title and the video duration is the average of the dataset.} 
  \label{table:2} 
  \begin{tabular}{ccccccc} 
    \toprule 
    {\bf Name} & {\bf Language} & {\bf Source} & \makecell{\bf Instances\\(Fake/Real)} & \makecell{\bf Title\\(Tokens)} & {\bf Duration} & {\bf Publisher} \\ 
    \midrule 
    FakeSV & Chinese & DY & 1827/1827 & 22.4 & 38.5s & Users \\
    FakeTT & English & TT & 1172/819 & 21.8 & 48.1s & Users \\ 
    \textbf{FMNV} & \textbf{English} & \textbf{YT\&TW} & \textbf{1500/893} & \textbf{21.9} & \textbf{73.8s} & \textbf{Media} \\
    \bottomrule 
  \end{tabular}
\end{table}

\subsection{Data Analysis}

The FMNV dataset comprises a total of 2,393 news videos, including 893 videos categorized as real and 1,500 videos categorized as fake. The fake category can be further divided into four subtypes: CD: 600 videos with title modifications causing semantic inconsistencies; CE: 450 videos where key video clips matching the title were deleted; SV: 300 videos with audio replaced by reports of different events un-der the same theme; CA: 150 videos featuring titles with synonymous but exaggerat-ed narration.
When compared to other MFND video datasets, FMNV offers notable advantages. In our LLM text generation process for data augmentation, we use a word substitu-tion strategy to replace some words from the original text to change the original meaning. This ensures that the length of both positive and negative texts remains comparable, thereby preventing significant length disparities. With an average video duration of 73.8 seconds, FMNV boasts considerably longer videos than other da-tasets, and nearly half of the videos are long videos longer than 80 seconds, allowing for a richer information content per video. A comparative analysis of MFND da-tasets is presented in Table~\ref{table:2}.

\section{Method}

Fig.~\ref{FNVD} illustrates the overall architecture of our proposed FMNVD model, com-prising two main components: feature extraction and feature aggregation. We use BERT to encode video titles and audio transcripts, and adopt a dual-stream framework combining 3D ResNeXt-101 and CLIP for dynamic and static visual fea-ture extraction. A co-attention-based aggregation module is then used to integrate multimodal features and perform fake news classification.

\subsection{Feature Extraction}

\subsubsection{Title} The video title is tokenized using the BERT-base WordPiece tokenizer with [CLS] and [SEP] tokens. The sequence is input to a 12-layer Transformer, and we extract the final hidden state of the [CLS] token as the title representation. To standardize input length, sequences are padded or truncated to $l_t=32$. The resulting token embeddings $F_t = [w_1, \dots, w_{l_t}]$ serve as contextualized text features.

\begin{figure*} \centering \includegraphics[width=\textwidth]{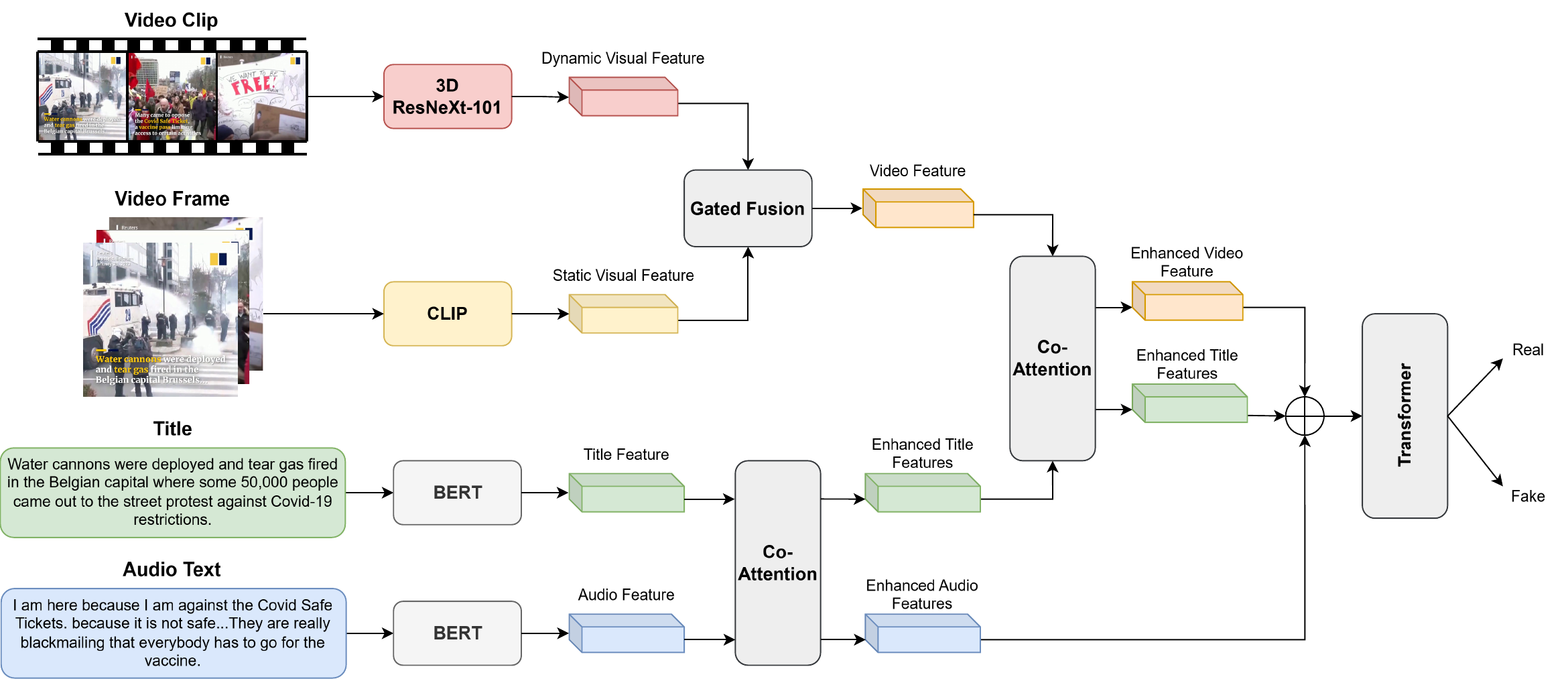} \caption{Overview of proposed FMNVD.} \label{FNVD} \end{figure*}

\subsubsection{Video}

We design a dual-stream feature extraction module to jointly capture temporal motion dynamics and static spatial semantics from video sequences. Given an input video $V$ of total duration $T$, we segment it into overlapping clips of 8 frames to extract motion features and sample individual keyframes for static analysis.

For motion representation, we utilize a pre-trained ResNeXt-101 model with 3D convolutions to encode spatio-temporal patterns from short clips. Let $V_{t:t+7}$ denote a segment of 8 consecutive frames; the motion feature is then defined as: 
\begin{equation}
F_m = \psi_m V_{t:t+7} \in R^{2048}
\end{equation}
where $\psi_m$ represents the 3D convolutional motion encoder.

In parallel, we extract static visual semantics using the CLIP visual encoder (ViT-B/32), which processes individual sampled frames. The resulting frame-level feature is expressed as:
\begin{equation}
F_s = \psi_s V_t \in R^{512}
\end{equation}
where $\psi_s$ denotes the CLIP visual encoder.

To enable fine-grained fusion of dynamic and static information, we introduce an attention-based gated fusion mechanism. Specifically, the motion and static features are linearly projected and concatenated, and a learnable sigmoid gate is applied to compute a dynamic weighting:
\begin{equation}
\alpha = \sigma\left(W[F_m; F_s] + b\right)
\end{equation}
\begin{equation}
F_v = \alpha F_m + (1 - \alpha) F_s
\end{equation}
Here, $W$ and $b$ are trainable parameters and $\sigma$ denotes the sigmoid function. The resulting fused visual embedding $F_v \in R^{d}$ encodes both short-term dynamics and long-range semantics.

\subsubsection{Audio} Audio is transcribed using Whisper and tokenized via BERT, following the title module. Sequences are padded/truncated to $l_a=64$, then encoded using the same BERT model. The first 64 token embeddings form $F_a = [w_1, \dots, w_{l_a}]$. Parameter sharing ensures alignment between title and audio representations.


\subsection{Feature Aggregation}

To integrate information across modalities, we adopt a hierarchical co-attention mechanism inspired by Qi et al.~\cite{DBLP:conf/aaai/0005BC0SXWC23}, which enables deep interactions among textual, auditory, and visual features.

We first apply bi-directional co-attention between title features $F_t \in {R}^{L_t \times d}$ and audio transcription features $F_a \in {R}^{L_a \times d}$, both projected into a shared space. Multi-head attention is computed in both directions:
\begin{equation}
\text{Attn}(Q, K, V) = \text{softmax}\left(\frac{QK^\top}{\sqrt{d_k}}\right)V
\end{equation}
with residual connections and layer normalization:
\begin{align}
F_t' &= \text{LayerNorm}(F_t + \text{Attn}{a \rightarrow t}) \\
F_a' &= \text{LayerNorm}(F_a + \text{Attn}{t \rightarrow a})
\end{align}

Next, the enhanced title features $F_t'$ attend to visual frame features $F_v \in {R}^{L_v \times d}$ using the same co-attention strategy, producing updated representations $F_t''$ and $F_v'$:
\begin{equation}
F_t'', F_v' = \text{CoAtt}(F_t', F_v)
\end{equation}

The final fused vector is formed by concatenating the three enhanced modalities:
\begin{equation}
F = \text{Concat}(F_t'', F_a', F_v') \in {R}^{3d}
\end{equation}
and passed through a Transformer encoder followed by an MLP for classification:
\begin{equation}
\hat{y} = \text{MLP}(\text{Transformer}(F W^\top))
\end{equation}

This hierarchical fusion enables the model to capture fine-grained cross-modal semantics and improves its robustness to subtle manipulations in fake news content.

\section{Experiments}


\subsection{Baseline}



1) \textbf{SV-FEND}~\cite{DBLP:conf/aaai/0005BC0SXWC23} The model leverages cross-modal attention mechanisms to enhance feature representation by capturing correlations between different modalities and integrates social context information  to improve detection accuracy. 2) \textbf{FakingRecipe}~\cite{DBLP:conf/mm/Bu000W024} This model detects fake news in short videos by analyzing material selection and editing behaviors. It captures sentiment and semantic clues from audio, text, and visuals during material selection, and spatial-temporal editing patterns during video editing, enhancing detection accuracy.

\subsection{Experimental Settings}

The models are optimized using the Adam optimizer~\cite{DBLP:journals/corr/KingmaB14} with an initial learning rate of \(1 \times 10^{-3}\). To ensure a rigorous evaluation, we measure four classification metrics: Accuracy, F1-score, Precision, and Recall. All experiments are conducted on an NVIDIA GeForce RTX 2080 Ti GPU with a batch size of 128 for 30 epochs. The cross-entropy loss function is minimized during training.


\subsection{Experiments Results}

\begin{table*}[t]
  \centering 
  \caption{Performance of different methods on FMNV.} 
  \label{table:3} 
  \begin{tabular}{cccccc} 
    \toprule 
    \textbf{Methods} & \textbf{Test Data} & \textbf{Accuracy} & \textbf{F1} & \textbf{Precision} & \textbf{Recall}\\
    \midrule 
    SV-FEND &
    \makecell{CD\\CE\\VS\\CA\\ALL} & 
    \makecell{60.00\\83.70\\82.50\\76.19\\72.92} &
    \makecell{55.44\\83.13\\80.43\\65.73\\72.26} &
    \makecell{56.94\\83.13\\79.41\\64.45\\72.26} &
    \makecell{55.83\\87.22\\88.33\\75.00\\73.67} \\
    \midrule 
    FakingRecipe & 
    \makecell{CD\\CE\\VS\\CA\\ALL} & 
    \makecell{60.67\\88.15\\88.33\\82.85\\73.75} &
    \makecell{52.88\\87.39\\86.32\\68.42\\72.70} &
    \makecell{57.27\\86.44\\84.09\\66.98\\72.16} &
    \makecell{54.72\\90.00\\92.22\\70.56\\72.77} \\
    \midrule 
    \textbf{FMNVD} & \makecell{CD\\CE\\VS\\CA\\ALL} & 
    \makecell{64.67\\87.41\\86.67\\77.14\\\textbf{74.17}} &
    \makecell{60.05\\86.75\\84.60\\61.44\\\textbf{73.76}} &
    \makecell{62.82\\86.00\\82.61\\60.34\\\textbf{74.17}} &
    \makecell{60.28\\90.00\\91.11\\64.44\\\textbf{75.78}}\\
    \bottomrule 
  \end{tabular}
\end{table*}

To validate the effectiveness of our proposed method and other baselines on the FMNV dataset, we conducted experiments by dividing the fake samples in the test set into four categories: Contextual Dishonesty (CD), Cherry-picked Editing (CE), Synthetic Voiceover (SV), and Cross-modal Attribution (CA), corresponding to four distinct manipulation strategies. Each model was evaluated on these subsets to analyze their detection performance across manipulation types. Additionally, we assessed performance on the full test set to examine overall effectiveness.

As summarized in Table~\ref{table:3}, all three models achieved over 0.72 accuracy, with our proposed FMNVD model attaining the highest score of 0.7417. This confirms both the generalization capability of the FMNV dataset and the competitive performance of FMNVD. Notably, the models perform well on CE and SV manipulations, while detection on CD samples remains challenging due to their subtle, phrase-level semantic alterations. This highlights a key limitation in current methods and points to future directions for improving fine-grained semantic reasoning.


\subsection{Ablation Study}

We further conduct a multimodal ablation study using FMNVD on the dataset to analyze the contributions of different modalities in FMNV. Specifically, we individually remove the title, video, and audio modalities from the data while simultaneously eliminating their corresponding cross-modal attention layers in the model. The experimental results presented in Table \ref{table:5} demonstrate that all three modality ablations caused notable performance degradation in the baseline model. The relative importance of different modalities to model performance can be ranked as audio \textgreater title \textgreater frames. This phenomenon primarily stems from our implementation strategy where audio signals are first converted into text through Whisper for processing. The transcribed audio text inherently contains more linguistic content than video titles, and such extended textual information proves more impactful than shorter textual inputs (titles) and visual features in our framework.

\begin{table}
  \centering 
  \caption{Results of ablation study.} 
  \label{table:5} 
  \begin{tabular}{ccccc} 
    \toprule 
    \textbf{Data} & \textbf{Accuracy} & \textbf{F1} & \textbf{Precision} & \textbf{Recall}\\ 
    \midrule 
    {All} & \textbf{74.17} & \textbf{73.76} & \textbf{74.17} & \textbf{75.78} \\
    \makecell{w/o Title\\w/o Frames\\w/o Audio} &
    \makecell{65.00\\72.50\\60.00} & 
    \makecell{64.00\\71.94\\57.87} &
    \makecell{64.06\\72.11\\57.81} &
    \makecell{64.89\\73.56\\58.00}\\
    \bottomrule 
  \end{tabular}
\end{table}

\section{Conclusion}

In this paper, we conduct an empirical analysis for fake news video categorization and construct a novel fake news video dataset (FMNV) using large model-based data augmentation. We provide a comprehensive description of FMNV's creation process with in-depth investigations into data entity attributes and distribution characteris-tics. To assess the dataset's validity, we introduce a new baseline model FMNVD and perform numerical experiments to evaluate various baseline performances, thereby demonstrating FMNV's generalization capability. Future work will explore additional promising approaches to enhance the dataset's predictive power, with the expectation that FMNV will pave the way for advanced developments in MFND research.

\end{document}